\documentclass[letterpaper]{article} 

\usepackage{aaai24}  
\usepackage{times}  
\usepackage{helvet}  
\usepackage{courier}  
\usepackage[hyphens]{url}  
\usepackage{graphicx} 
\usepackage{natbib}  
\usepackage{caption} 
\usepackage{amsmath}
\usepackage[pagebackref]{hyperref}

\usepackage{enumitem}
\usepackage{booktabs}
\usepackage{multirow}
\usepackage{algorithm}
\usepackage{algpseudocode}
\usepackage{amsfonts}
\usepackage{makecell}
\usepackage[table]{xcolor}

\usepackage{newfloat}
\usepackage{listings}
\nocopyright
\DeclareCaptionStyle{ruled}{labelfont=normalfont,labelsep=colon,strut=off} 
\floatstyle{ruled}
\newfloat{listing}{tb}{lst}{}
\floatname{listing}{Listing}
\title{Embracing the black box: \break Heading towards foundation models for causal discovery from time series data}
\author {
    Gideon Stein,
    Maha Shadaydeh,
    Joachim Denzler
}
\affiliations {
      Computer Vision Group, Friedrich Schiller University Jena\\
        gideon.stein@uni-jena.de, maha.shadaydeh@uni-jena.de, joachim.denzler@uni-jena.de
}

\begin{document}

\maketitle

\begin{abstract}
Causal discovery from time series data encompasses many existing solutions, including those based on deep learning techniques. However, these methods typically do not endorse one of the most prevalent paradigms in deep learning: End-to-end learning. To address this gap, 
we explore what we call Causal Pretraining.
A methodology that aims to learn a direct mapping from multivariate time series to the underlying causal graphs in a supervised manner.
Our empirical findings suggest that causal discovery in a supervised manner is possible, assuming that the training and test time series samples share most of their dynamics. More importantly, 
 we found evidence that the performance of Causal Pretraining can increase with data and model size, even if the additional data do not share the same dynamics. Further, we provide examples where causal discovery for real-world data with causally pretrained neural networks is possible within limits. We argue that this hints at the possibility of a foundation model for causal discovery.
\end{abstract}

\section{Introduction}

\begin{figure}[ht!]
\centering
\includegraphics[width=0.99\linewidth]{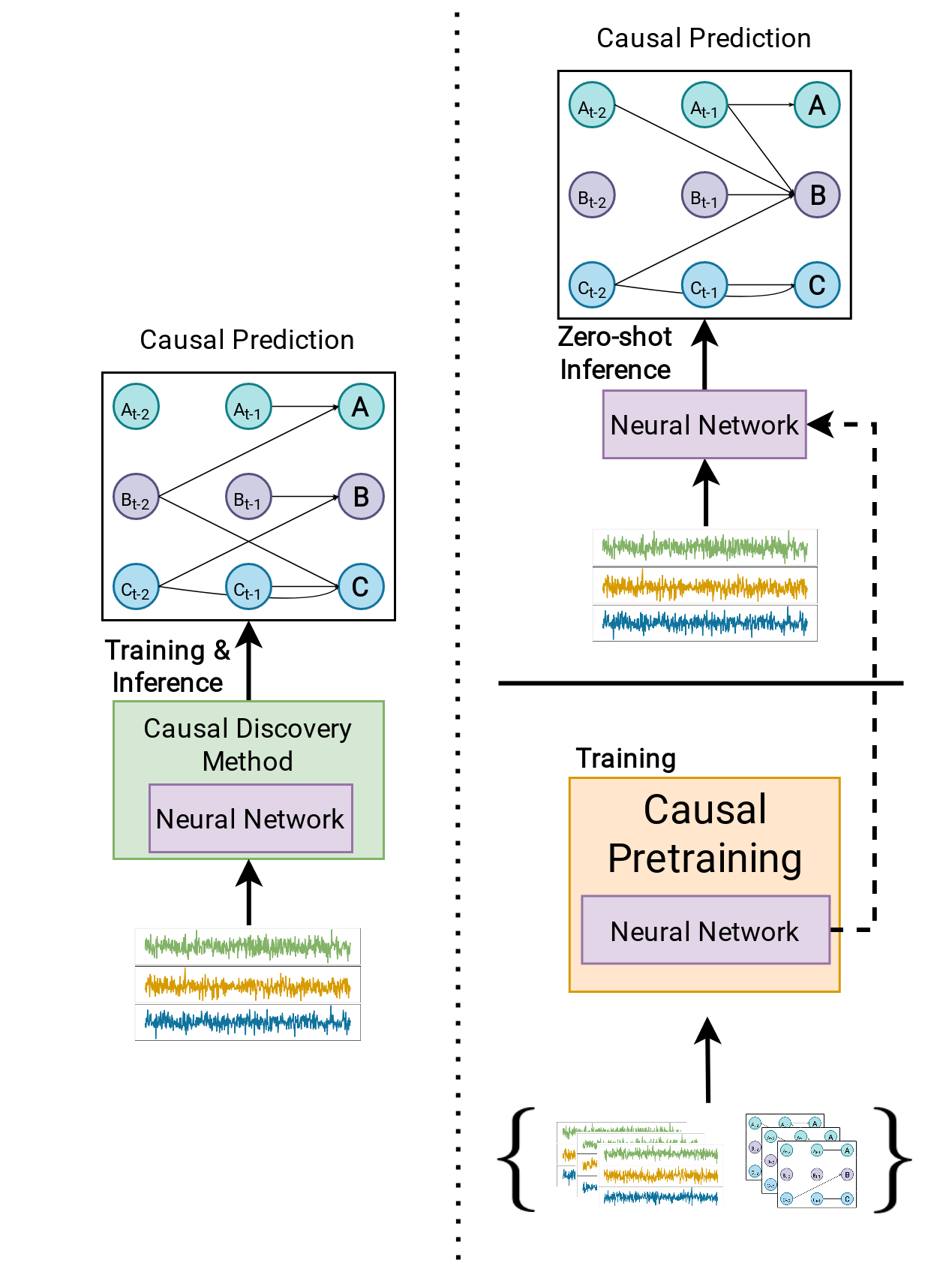}
\caption{Comparison between a general depiction of causal discovery methods (left) and our Causal Pretraining methodology (right). Instead of inferring causal graphs from data directly, Causal Pretraining produces neural networks that can be deployed for inference directly.}
\label{g:1}
\end{figure}


The investigation of causal discovery from time series data encompasses many existing solutions, including, of course, the integration of deep learning techniques.
Despite the widespread utilization of these techniques, many methods do not endorse one of the most prevalent paradigms in deep learning: End-to-end learning.
This paradigm suggests that it is often beneficial to only provide neural networks with raw input data instead of handcrafting features. Additionally, making assumptions on how a particular task should be solved can hinder performance \cite{bojarski_end_2016}, \cite{amodei_deep_2016}, \cite{glasmachers_limits_2017}.
Contrary to that, neural networks are typically embedded into well-established causal discovery frameworks such as Granger causality, \cite{tank_interpretable_2018}, \cite{ahmad_causal_2022}, \cite{teodora_trifunov_time_2022}, \cite{lowe_amortized_2022} or score-based methods \cite{zheng_dags_2018}, \cite{ng_graph_2019}. Through this, the solution space is naturally restricted.
To address this disparity, we explore what we call \textbf{Causal Pretraining}, a methodology that aims at learning a direct mapping from multivariate time series to causal graphs as depicted in \autoref{g:1}. We propose to parameterize this mapping by a deep neural network and train it in a supervised manner. In the training phase, we sample from a distribution of synthetic time series with known corresponding causal graphs.
Besides leaving the solution space completely open, Causal Pretraining has another distinct advantage over many other deep-learning approaches. Since it does not require fitting any parameters during inference and natively allows for effective parallel processing, it makes the analysis of large sets of time series exceptionally efficient.
While this idea is, to our surprise, largely unexplored,  we determined that substantial challenges are stabilizing training and encouraging behavior beyond learning correlational patterns. 
Therefore, we conducted experiments involving training multiple deep-learning architecture archetypes with differing training techniques on increasingly more complex synthetic datasets. 
We specifically evaluate the test performance for synthetic data samples that mimic the training dynamics (in-distribution) and synthetic data samples with slightly altered dynamics (out-of-distribution). To demonstrate the applicability of our approach to real-world data, we also test our approach on in-the-wild datasets with completely unknown data distributions in a zero-shot setting.

While much work still lies ahead, we empirically show that Causal Pretraining can consistently uncover causal relationships for unseen time series, assuming that the training and test time series samples share most of their dynamics. More importantly, we also find that the performance of Causal Pretraining increases with model size and the amount of training data, even when the additional training samples extend the data distribution. In these scenarios, we show that Causal Pretraining can outperform alternative simple approaches without fitting parameters for new data. We further find potential to truly extrapolate to real-life data. We argue that this together hints at the possibility of a foundation model for causal discovery. Finally, we emphasize that this paper focuses on empirical results and the applicability of CP to real-world data. We retain theoretical insights, e.g., identifiability statements, for future work.
In summary, we present  the following contributions:
\begin{enumerate}[label=\textbf{\arabic*}.]

\item  We explore supervised learning for causal discovery from time series data, aiming at learning it end-to-end from synthetic data with shared dynamical properties. 
\item We introduce and evaluate several helper techniques to support the performance of Causal Pretraining. 
\item We evaluate the ability of causally pretrained neural networks to predict real-world causal graphs.
\item We observe relations to foundation models, as supported by our empirical results.

\end{enumerate}

\section{Background and Related Work}
The discovery of causal relationships from observational time series can be tackled in various ways \cite{gong_causal_2023},  \cite{assaad_survey_2022}, \cite{vowels_dx2019ya_2022}, \cite{runge_causal_2023}. 
Here, we give an overview of these methods and elaborate on how they relate to causal pretraining. 

Score-based approaches aim to find a causal graph that maximizes a defined scoring function (e.g., AIC score \cite{kotz_information_1992}).
With the introduction of formulating the scoring as a continuous optimization problem 
\cite{zheng_dags_2018}, \cite{pamfil_dynotears_2020}, the possibility of deploying neural networks arose quickly. 
For causal discovery in data without temporal dimension (sample data), a multitude of methods that deploy deep-learning architectures was developed (\cite{lachapelle_gradient-based_2020}, \cite{ng_graph_2019},\cite{kyono_castle_2020}).
Concerning time series data, \cite{sun_nts-notears_2023} adapts the previously used vectorized autoregressive model (VAR) from \cite{pamfil_dynotears_2020} to handle nonlinear relationships by deploying a 1D convolution-based architecture.

Being exclusively applicable to time-series data, Granger causality\cite{granger_investigating_1969} has a substantial history.
At its core, it evaluates whether a certain variable's history helps with predicting another target variable. Deep learning-based Granger causality methods mostly deploy neural networks as forecasting architectures and infer Granger causality in various ways. \cite{montalto_neural_2015} and its extension \cite{wang_estimating_2018} deploy a greedy search over which input variable to use with neural networks to determine
Granger-causes. \cite{tank_interpretable_2018} and \cite{tank_neural_2021} alternatively use the weights of the first layer of an MLP under $L2$ regularization of the same layer to determine Granger-causes. Other methods infer from attention weights \cite{guo_exploring_2019}, \cite{dang_seq2graph_2018} or the parametrization of variational autoencoders \cite{li_causal_2023}, \cite{meng_estimating_2019}.
Furthermore, approaches such as \cite{ahmad_causal_2022} propose to generate in-distribution intervention variables and determine Granger causes using the model invariance assumption. Importantly, in all approaches mentioned so far, neural networks are optimized for a single specific dataset.

Framing causal discovery as a supervised learning task was also explored to some extent. 
Concerning sample data, an early work that does not deploy deep learning is \cite{lopez-paz_towards_2015}, which aims at learning a binary classifier to direct a link between two variables from data.
Some recent strain work that leverages deep learning methods to map from data correlation matrix to 
directed acyclic graph is \cite{li_supervised_2020} and \cite{petersen_causal_2023}.
\cite{geffner_deep_2022} aims at learning an autoregressive flow model \cite{huang_neural_2018} from data that can be used for causal discovery.
Finally, as the work that is closest to our methodology, \cite{lowe_amortized_2022} aims at learning a mapping from multivariate time series to causal graph by deploying a Variational autoencoder-based architecture to forecast the time series.
The encoder output is then leveraged to infer Granger-causes for each variable. On the contrary, we aim to learn direct mappings without a surrogate task. Another crucial disparity from \cite{lowe_amortized_2022} is that we frame Causal Pretraining to uncover the time lags of any causal relationship rather than only returning a summary graph that ignores temporal patterns.
To the best of our knowledge, our proposed method is the first that aims to learn causal discovery for time series completely end-to-end.

\section{Method}
\label{sec:method}
\begin{figure*}[ht!]
\centering
\includegraphics[width=0.9\linewidth]{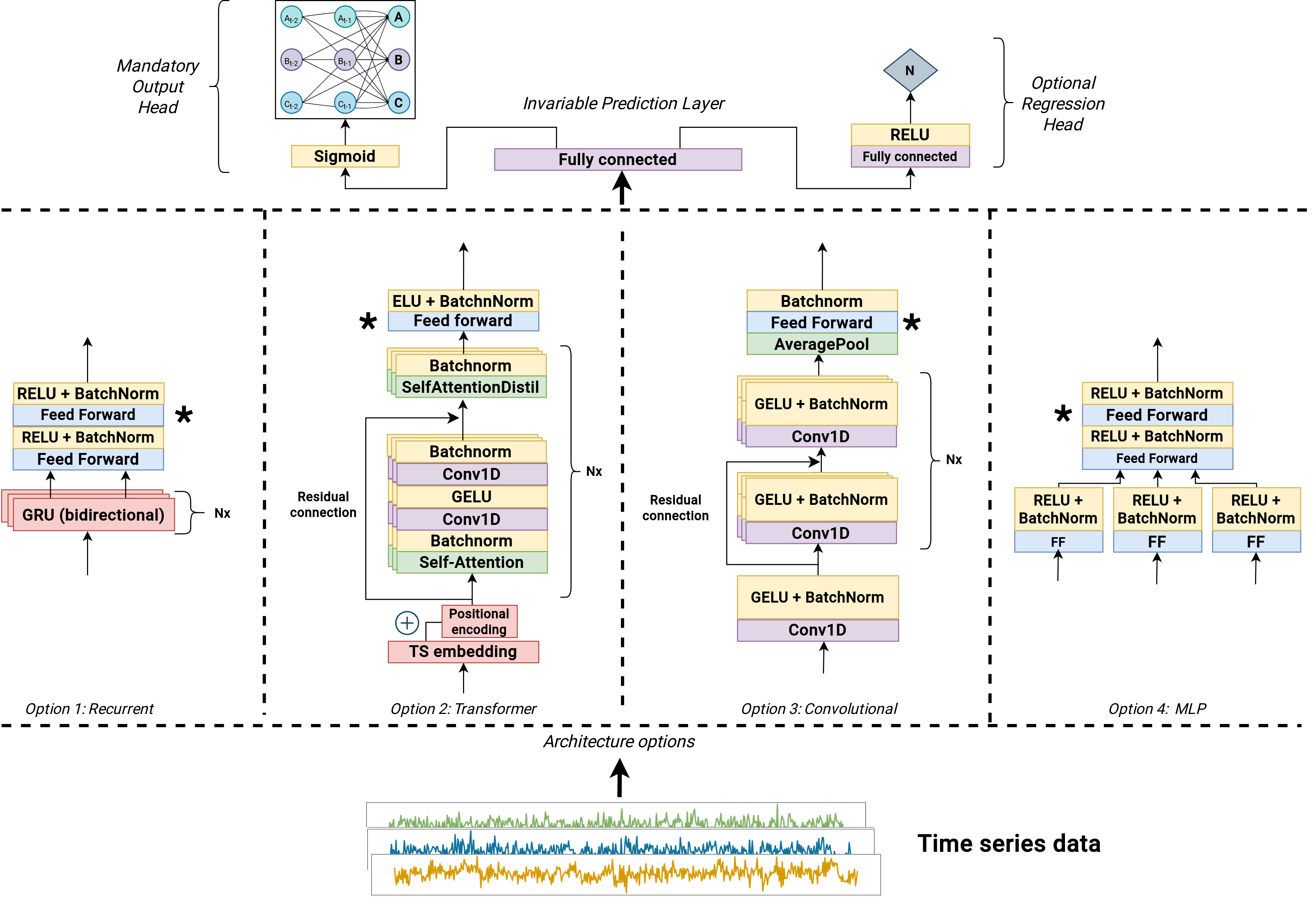}
\caption{Depiction of architectures that we consider for Causal Pretraining. From left to right: \textbf{GRU, Transformer, ConvMixer, MLP}. We further mark locations for correlation injection with \textbf{*}.}
\label{g:2}
\end{figure*}

Here, we formally introduce our methodology and discuss assumptions. We subsequently detail the architectures and the data generation process we used to test our approach. Finally, we suggest various training techniques and assess their impact on Causal Pretraining.

\subsection{Definitions \& Assumptions}
 We define $\mathbb{X}= (x^t_i)_{i=1, .., V; t=0, .., T}$ to be a multivariate time series with $V{ \in \mathbb{N}^+}$ variables and a length of $T{ \in \mathbb{N}_0}$.
We assume, for now, that any $x_i^t$ contained in $\mathbb{X}$ can be described by the following time-invariant structural equation model:

\begin{equation}
\label{eq:1}
x_i^{t} = \sum_{n=1}^{N} \sum_{j=1}^{V} f_{i,j}^{n}(\mathbb{A}^{i,j, n}\cdot x^{t-n}_j)  + e^t_i ~~~~.
\end{equation}
Here $\mathbb{A}$ is a three-dimensional tensor where element $\mathbb{A}^{j,i,n} $ denotes a coefficient that describes the (causal) impact of a specific variable $j$ at a specific time lag $n$,  i.e., $x^{t-n}_j$,  on $x^t_i$. $N$ denotes the maximum time lag considered. Each linear effect is then transformed by an individual function $ f_{j,i}^{n}$ to account for possible nonlinear relationships. 
 Additionally, $(e^t_i)_{i=1, .., V}$ denotes uncorrelated Gaussian noise. 

Causal discovery methods (CDMs) are concerned with uncovering the set of non-zero elements of $\mathbb{A}$ from a given time series $\mathbb{X}$.
Typically they either aim at uncovering a summary graph that is represented by an adjacency matrix $G \in \mathbb{Z}^{V} \times V$ where element $(i,j)$ denotes whether any current or past observation of variable $x_j$ has an impact on variable $x^i$. Formulated differently, it specifies whether the parameter vector $\mathbb{A}^{i,j}$ has any non-zero elements.
Alternatively, methods that aim at uncovering a window causal graph return a 3-dimensional tensor $G \in \mathbb{Z}^{V \times V \times N}$ where element $(i,j,n)$ specifies whether $\mathbb{A}^{i,j,n}$ is estimated to be non-zero.
Based on this formalism, we can naively summarize CDMs as: $\textrm{CDM}(\mathbb{X}) \rightarrow G$.
Contrary to that, Causal Pretraining is not concerned with directly uncovering any causal graph $G$, but rather with producing a causally pretrained neural network (CPNN) in a supervised manner that can be applied in zero-shot settings (no further parameter optimization) to infer $G$. 
We define $\mathcal{G}$ as a set of structural equations (\autoref{eq:1}) that each specifies the causal dynamic of a multivariate time series.
Additionally, we define $\mathcal{X}$ to be a set of corresponding $\mathbb{X}$ that originates from a specific element in $\mathcal{G}$.
Formally, Causal Pretraining (CP) can be defined as follows:

\begin{align}
\textrm{CP}(\mathcal{X},\mathcal{G})  \mapsto \textrm{CPNN} , \hspace{0.25cm}
\textrm{CPNN}(\mathbb{X}) \mapsto G \in \mathbb{Z}^{V \times V \times N},
\label{eq:3}
\end{align}
 where CPNN  refers to a causally pretrained neural network. We treat the elements of the network output $G$ as probability estimates that the corresponding elements in $\mathbb{A}$ exist. Throughout this paper, we make the following assumptions: 1. $G$ is a Directed Acyclic Graph (DAG). No instantaneous effects. 2. Causal sufficiency: All causes are observed. 3. Theoretical insights such as \cite{shah_hardness_2020} suggest that learning a CPNN that can correctly identify the non-zero elements of $\mathbb{A}$ for any $\mathbb{X}$ is very likely impossible. Instead, we assume that train and test distribution share general dynamics, such as the functional space $f$ in \autoref{eq:1} from which we draw elements for all synthetic data samples.


\subsection{Architectures}
To properly evaluate our methodology, we deem it essential to evaluate multiple differing architectures. 
We, therefore, select five different architectures, aiming at covering the prominent deep-learning archetypes up to this date and not necessarily focusing on fully optimizing the architecture for Causal Pretraining for now. We test an \textbf{MLP} \cite{taud_multilayer_2018}, a unidirectional GRU \textbf{(uGRU)}, a bidirectional GRU (\textbf{bGRU}), a Conv Mixer \textbf{CM} \cite{trockman_patches_2022},  a recently introduced convolutional architecture that showed strong results on image tasks, which we adapt for time series and a Transformer  \textbf{(Trf)} that borrows heavily from the Informer \cite{zhou_informer_2021} architecture excluding sparse attention but featuring Attention Distillation \cite{zhou_informer_2021}. All architectures are depicted in \autoref{g:2} (both GRU architectures are jointly depicted).
We keep input and output dimensions for all architectures the same (input: $\mathbb{X}$, output: causal graph $G$, optional: regression surrogate task as described below).
All networks return a vector with the length $V * V * N$ that we interpret as probabilities for specific elements of $\mathbb{A}$ to be non-zero.
We reshape this vector into $G^{V \times V \times N }$ and then calculate binary cross-entropy for the primary loss component with $\mathbb{A} > 0$ as the label.
To make the comparison fair, we scale the architectures to have similar amounts of parameters, aiming at estimating the innate capabilities of these archetypes. In total, we deploy five sizes in our experiments (although not all sizes during every experiment),  which we denote as "small" ($\sim$13k params), "medium" ($\sim$120k params), and "big" ($\sim$1.5M params) "deep" ($\sim$17M params), and "lcm" ("large causal model", $\sim$~300M params), see\hyperref[app:3]{Appendix~\ref*{app:3}}.

\subsection{Additional Training Techniques}
While deep learning architectures are, in theory, universal function approximations \cite{hornik_multilayer_1989}, in reality, learning a proper function from observational data is not always straightforward.
Networks often converge to local minima or memorize data, which we attempt to counter by employing the following three techniques to help with the optimization process.

\paragraph{Regression Output}
An easy-to-achieve suboptimal solution in our training setup is predicting all elements of $\mathbb{A}$ as zero. This often results in a small loss given that the true number of non-zero elements in $\mathbb{A}$ ($|\mathbb{A} > 0|$) is small.
To discourage this solution, we add an optional regression output that predicts the number of non-zero elements in $\mathbb{A}$. We formulate the loss related to this prediction as follows: 
\begin{equation}
\label{eq:5}
\textrm{regLoss} = (\hat{Y}_{reg} - |(\mathbb{A} > 0)|)^{2}
\end{equation}
where $\hat{Y}_{reg}$ specifies the regression output.
This task should be trivial when the correct edges are identified (counting), and this penalty term should be close to zero. However, predicting a graph with no edges leads to a high penalty. 
\paragraph{Correlation Regularization}

Additionally, since no correlation typically implies no causation, we introduce a regularization term called correlation regularization (CR). This term punishes the case where a model is confident that a specific element of $\mathbb{A}$ is non-zero (high confidence in the model output $G$), but the lagged-crosscorrelation of the corresponding time series is low. CR is defined in the following way:



\begin{align}
CR(G, \mathbb{X}) &= \sum_{i=1}^{V} \sum_{j=1}^{V} \sum_{n=1}^{N} \left(
\frac{G^{i,j,n}}{|\mathit{lcc}(\mathbb{X}_i, \mathbb{X}_j, n )  | + \beta}
\right) ^\alpha , \\
\mathit{lcc}(a,b,n) &= \frac{\sum\limits_{t=0}^T(a^t - \bar{a})  \cdot L^n(b^t - \bar{b})}{\sqrt{\sum\limits_{t=0}^T(a^t- \bar{a})^2}  \sqrt{\sum\limits_{t=0}^T \left(L^n(b^t - \bar{b})\right)^2 }},
\end{align}
where $L^n$ is the lag operator (shifting the time series $n$ steps).
Furthermore, $\alpha$ and $\beta$ are hyperparameters determining the exact shape of the function.
We include a depiction of CR in the \hyperref[g:3]{Appendix~\ref*{g:3}}.
Notable, the term does not reward high confidence for predictions with corresponding high cross-correlation. Nor does it punish low confidence for edges with corresponding high cross-correlation values. It thereby only encodes the idea that no correlation implies no causation.

\paragraph{Correlation Injection}

Many causal discovery methods take a skeleton graph that encodes correlations between variables as a starting point for causal discovery (\cite{spirtes_causation_2000},\cite{li_supervised_2020},\cite{petersen_causal_2023}). To mimic this and to help the learning process, we fuse (concatenate to the current hidden state) the lagged cross-correlation of all time series in $\mathbb{X}$ into the network as depicted in \autoref{g:2}.

\subsection{Data}
\label{data:1}
\subsubsection{Synthetic Data}
To perform Causal Pretraining, we require a large set of training examples. As real-world data with labels, i.e., ground-truth causal graphs, are typically rare, we generate sets of synthetic data ($\mathcal{G}, \mathcal{X}$) to train on. 
We randomly select elements of $\mathbb{A}$ to be non-zero to create a single training example.
We then randomly sample all values of $\mathbb{A}$ that are non-zero from a specified uniform distribution and draw nonlinear relations $f_{i,j}^n$ from a set of specified functions:\{$e^x$, $x^2$, $\sigma(x)$, $sin(x)$, $cos(x)$, $relu(x)$, $log(\sigma(x))$,  $\frac{1}{x}$, $\|x\|$, $clamp(x,(-0.5,0.5)$\}.
For linear datasets, we generate data from vector autoregressive models, i.e., we set all $f^n_{i,j}(\cdot)=(\cdot)$ in \autoref{eq:1}.
Then we generate $\mathbb{X}$ according to this dynamic.
A detailed description of this process is included in \hyperref[app:1]{Appendix~\ref*{app:1}}.
Since there is no guarantee that randomly sampled structural equations (\autoref{eq:1}) lead to a stable system, we establish stability tests to exclude unstable systems from the datasets. 
Finally, we apply min-max normalization to the time series. We refer to \hyperref[app:2]{Table~\ref*{app:2} (Appendix)} 
for further specifications of the datasets that we use for training. We also provide additional data information in the \hyperref[experiments:1]{Experiment section~\ref*{experiments:1}}.

\subsubsection{Kuramoto}
A frequently used source for synthetic data is fully observable physical simulations. Similar to \cite{lowe_amortized_2022} from which we also adapt the simulation,  we perform Causal Pretraining on data originating from a Kuramoto model \cite{kuramoto_self-entrainment_1975} with five variables. Since we are primarily concerned with one-dimensional variables, we use the trajectories of the simulated variables as one-dimensional time series. Notably, the dynamics of this dataset do not strictly follow \autoref{eq:1}. We, therefore, empirically evaluate with this dataset whether we can relax our initial assumption.

\subsubsection{Zero-shot Inference}
Another capability of Causal Pretraining we aim to explore in this work is whether it can extrapolate to real-world data outside the training distribution. To test this, we perform zero-shot inference on two distinct benchmarks, meaning we do not perform any further parameter optimization for unseen data.
First, we predict the causal relationships in a benchmark involving the discharge of rivers in Germany \cite{ahmad_causal_2022}. 
Secondly, we test the causal direction in a real-life dataset originating from \cite{jesson_using_2021}, involving the estimation of aerosol-cloud interactions. 
Since the dataset is originally concerned with estimating the average treatment effect of aerosol on different cloud parameters under weather confounding (meaning the direction of the effect is clear), we are concerned with confirming this causal direction by only evaluating the predicted direction between these two variables under the influence of other weather variables (we consider sea surface temperature, effective inversion strength, and relative humidity at 850 mb).

\section{Experiments}
\label{experiments:1}
To properly evaluate Causal Pretraining, we conducted four studies aiming at the step-by-step evaluation. During all experiments, we compare CPNNs with two simple baseline techniques and one popular causal discovery method for time series data (PCMCI, \cite{runge_detecting_2019}). We additionally compare the results on the Kuramoto data with results provided in \cite{lowe_amortized_2022}.
For baselines, we calculate the absolute lagged cross-correlation for all variables and use them directly as probability estimates for $\mathbb{A}$. We refer to this as Correlation Thresholding (\textbf{CT}).
Additionally, we perform a linear form of Granger-causality \textbf{GVAR} by fitting a VAR model to the data and treating the absolute parameters of the model as probability estimates for $\mathbb{A}$.
Further, we deploy the \textbf{PCMCI} algorithm using partial correlation as the conditional independence test. 

\begin{table*}[ht]
    \centering
    
    \begin{tabular}{l|lccccc|ccc}
    
    \toprule
    
        ~ & ~& \textbf{MLP} &\textbf{uGRU}  & \textbf{bGRU} & \textbf{CM} & \textbf{Trf}& \textbf{CT}& \textbf{GVAR} & \textbf{PCMCI}\\ \hline
        ~ & ~& \multicolumn{5}{c|}{\textit{Causal Pretraining}}& \multicolumn{3}{c}{\textit{Baselines}}\\ \hline
        \rowcolor[HTML]{82B366}
        \multicolumn{10}{c}{\textit{Test-Set 1}}\\ 
        
        \midrule
        \rowcolor[HTML]{eff6ee} ~ & \textbf{SL} & .999 (.000) & \textbf{1.00 (.000)} & .999 (.000) & .999 (.000) & \textbf{1.00 (.000)} & .997 & \textbf{1.00} & \textbf{1.00} \\
        \rowcolor[HTML]{eff6ee}~ & \textbf{ML} & .498 (.005) & .621 (.053) & .626 (.185) & .513 (.014) & .621 (.125) & .997 &\textbf{ 1.00 }& \textbf{1.00} \\
        \rowcolor[HTML]{eff6ee}~ & \textbf{SNL} & .950 (.001) & \textbf{.955 (.001)} & .948 (.002) & .949 (.002) & .955 (.003) & .918 & .916 & .915 \\
        \rowcolor[HTML]{eff6ee}~ & \textbf{MNL} & .525 (.004) & .829 (.102) & .555 (.104) & .519 (.004) & .545 (.009) & .937 & .938 & \textbf{.941} \\
        \rowcolor[HTML]{eff6ee}~ & \textbf{LNL} & .933 (.001) & .937 (.001) & .937 (.001) & .935 (.002) & .934 (.001) & .936 & \textbf{.944 }& .943 \\
        \rowcolor[HTML]{eff6ee}  \multirow{-7}{*}         {\rotatebox[origin=c]{90}{\textbf{Single}}} & \textbf{XLNL} & .499 (.006) & .689 (.111) & .749 (.138) & .506 (.004) & .518 (.034) & .938 & \textbf{.944 }& .943 \\
        \midrule
        \midrule
        \rowcolor[HTML]{eff6ee}~ & \textbf{Medium}  &.928 (.023) & .907 (.023) & .920 (.012) & .887 (.003) & .950 (.015) & ~ & ~ & ~\\
        \rowcolor[HTML]{eff6ee}  \multirow{-2}{*}         {\rotatebox[origin=c]{90}{\textbf{Joint}}}& \textbf{Big} &.761 (.028) & .959 (.028) & .962 (.032) & .886 (.003) & \textbf{.977 (.001)} & \multirow{-2}{*}{.958} & \multirow{ -2}{*}{.888} & \multirow{-2}{*}{.946} \\

        \midrule
        \rowcolor[HTML]{D79B00}
        \multicolumn{10}{c}{\textit{Test-Set 2}}\\ 

        \midrule
        \rowcolor[HTML]{fff8e6}~ & \textbf{SL} & .999 (.000) &\textbf{ 1.00 (.000) }& .999 (.000) & .999 (.000) & \textbf{1.00 (.000) }& .999 & .999 & .999 \\
        \rowcolor[HTML]{fff8e6}~ & \textbf{ML} & .499 (.003) & .600 (.052) & .613 (.181) & .506 (.007) & .601 (.119) & .999 & \textbf{1.00} & 1.00 \\
        \rowcolor[HTML]{fff8e6}~ & \textbf{SNL} & .930 (.001) & \textbf{.935 (.000)} & .933 (.001) & .931 (.002) & .935 (.002) & .926 & .928 & .924 \\
        \rowcolor[HTML]{fff8e6}~ & \textbf{MNL} & .520 (.006) & .785 (.088) & .547 (.090) & .518 (.005) & .538 (.010) & \textbf{.927 }&\textbf{ .927} & \textbf{.927 }\\
        \rowcolor[HTML]{fff8e6}~ & \textbf{LNL} & .905 (.002) & .915 (.002) & \textbf{.916 (.001)} & .912 (.002) & \textbf{.916 (.001)} & .910 & .912 & .905 \\
        \rowcolor[HTML]{fff8e6}  \multirow{-6}{*}         {\rotatebox[origin=c]{90}{\textbf{Single}}} & \textbf{XLNL} & .502 (.006) & .659 (.096) & .710 (.124) & .503 (.006) & .518 (.027) & \textbf{.913 }& \textbf{.913 }& .910 \\
        \midrule\midrule

        \rowcolor[HTML]{fff8e6}~ & \textbf{Medium}& .916 (.023) & .895 (.022) & .906 (.011) & .869 (.003) & .939 (.015)& ~ & ~ & ~ \\
        \rowcolor[HTML]{fff8e6}  \multirow{-2}{*}         {\rotatebox[origin=c]{90}{\textbf{Joint}}} & \textbf{Big} & .752 (.027) & .946 (.038) & .952 (.033) & .867 (.002) &\textbf{.970 (.001)} & \multirow{-2}{*}{.948} & \multirow{-2}{*}{.862} &\multirow{-2}{*}{.910} \\

    \bottomrule
    
    \end{tabular}
  \caption{Mean AUROC scores for the synthetic data experiments \textbf{Single} and \textbf{Joint} (bottom two lines of each Test-set). For experiment \textbf{Single}, the first column specifics the dataset, while for experiment \textbf{Joint}, it specifies model size. Best results are denoted in \textbf{bold}. We report the corresponding standard deviation calculated over 10 runs in brackets. }
\label{tab:1}
\end{table*}

    
    
    






\subsection{Synthetic Data 1 - Impact of the Dataset}
\label{synth:1}
To evaluate the performance of CP, we conducted experiments on the precise data distribution we described in the \hyperref[data:1]{Data section~\ref*{data:1}}.
We continuously evaluated two distinct test sets to understand extrapolation capabilities during this synthetic data experiment. Firstly, a test set, named \textit{Test-Set 1}, includes test samples originating from the same distribution as the training data. Secondly, a test set, named  \textit{Test-Set 2}, which has an increased variance for the noise $e$ and an altered (slightly lower, not overlapping) range from which we draw non-zero elements in $\mathbb{A}$ (see \autoref{eq:1}).
To stop training, we perform early stopping based on the loss of an additional validation set that follows the $\mathbb{A}$ range of \textit{Test-Set 2} but keeps the training variance of the noise $e$. We use AdamW \cite{loshchilov_decoupled_2018} as the optimizer during all experiments.

We trained all described architectures on six distinct datasets. Here, we generated 5000 samples for training and 500 samples for testing, which have different structural equations (\autoref{eq:1}) with varying $\mathbb{A}$ and differing $f$.
We increased the complexity of the dataset step by step, increasing the number of variables, the number of lags, the coefficient range, and the set from which we draw the functions $f_{i,j}^n$.
In short, we tested two linear datasets \textbf{(SL, ML)}, two nonlinear datasets with a reduced nonlinear function set \textbf{(SNL, MNL)}, and two datasets with the full function set, as described in the \hyperref[data:1]{Data section~\ref*{data:1}},\textbf{(LNL, XLNL)}. 
In all three cases, we considered three variables, a maximum lag of two, and the model size "small" for the first dataset. For the second dataset, we considered five variables, a maximum lag of three, and the model size "medium".
To compare the performance of our method, we report the best-performing hyperparameter combination of each architecture. We 
searched the optimal parameters of batch size, learning rate, weight decay, and training addition (Regression Loss, CR, Correlation injection) by performing a full grid search 
and ran each hyperparameter combination twice.
We then selected the combination that resulted in the lowest observed validation loss and rerun it ten times.

Since we were also interested in the performance of CP when trained on data that origins from different distributions, we conducted a second set of experiments where we doubled the size of the six training sets that we described above and joined them together, making the maximum number of lags and the number of variables flexible. By this, we also mixed linear and non-linear samples, effectively drawing from different data distributions. We performed a similar hyperparameter search as in our first experiment. To keep the search space reasonable, we omitted the regression head. We also always included correlation injection since these hyperparameter selections performed consistently better in our first synthetic data experiment. We independently trained model sizes 'medium' and 'big' and rerun the best-scoring hyperparameter combination ten times.
We report the results of these studies in \autoref{tab:1}. We denote the results of the first study with \textbf{Single} and the results of the second study with \textbf{Joint}.

Surprisingly, we found that CP generally performed worse than our baselines in our first set of experiments. With some exceptions, the AUROC scores on both test sets were typically worse, even considering the best possible runs we performed. From our architectures, uGRU and Trf seemed to perform slightly better. However, these tendencies were too inconsistent to conclude a clear trend. 
Contrary to that, the performance of CP in our second experiment set outperformed all baselines, notably, while fitting no parameters on the test sets. Since CDNNs have to learn a much more extensive training distribution to perform comparably, we expected CDNNs to be less precise. We found the opposite to be true. Even if we kept the architecture size the same, e.g., 'medium', we observed that the performance of CDNNs increased. Furthermore, increasing the model size improved the performance and, most importantly, the generalization (Test-Set 2).
We interpret this in the following way: The broader the training distribution, the more robust/generally applicable CPNNs must become to solve the training distribution. This, in turn, helps with out-of-training performance and generalization. Further, the amount of samples required to optimize CP properly is much larger than we initially expected. We suggest that further improvements could be achieved by making the training distribution even broader, even for similarly sized architectures.

 \subsection{Synthetic Data 2 - Impact of Model Size}
\label{synth:2}

\begin{figure}[{h}]
\includegraphics[width=0.88\linewidth]{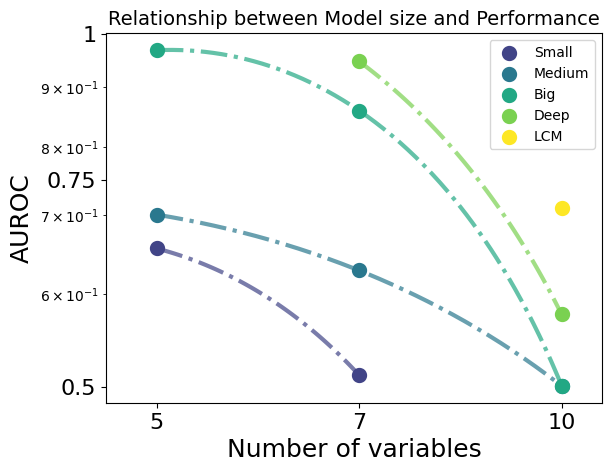}
\caption{Relationship between the Model size and the performance on datasets with an increased number of variables. We display the performance on Test-set 2 for each data point.}
\label{g:4}
\end{figure}

\begin{table*}
    \centering
    
    \begin{tabular}{lc|ccccc|ccc}

    \toprule
    
        ~ & \textbf{ACD*} & \textbf{MLP} &\textbf{uGRU}  & \textbf{bGRU} & \textbf{CM} & \textbf{Trf}& \textbf{CT}& \textbf{GVAR} & \textbf{PCMCI}\\
        \rowcolor[HTML]{82B366}

        \midrule
         \rowcolor[HTML]{80bcff}\multicolumn{10}{c}{\textit{Test-Set}}\\
        \midrule

            \rowcolor[HTML]{EFF6FF}\textbf{Kuramoto} & .952 &  .825 (.025) & .905 (.053) & .852 (.076) & \textbf{.954 (.010)} & .888 (.070) & 0.628 & 0.464 & 0.523  \\
\midrule
     \rowcolor[HTML]{EFF6FF}\textbf{Rivers} & ~  &0.425 & 1.00 & 0.5000 & 0.600 & 1.000 & 1.00 & 1.00 & 1.00 \\
        \rowcolor[HTML]{EFF6FF}\textbf{Aerosol-Cloud} & ~ &0.00 & 0.017 & 0.793 & 1.00 & 0.150 & 0.473& 1.00 & 0.363 \\

        \bottomrule
    \end{tabular}
  \caption{We here report mean AUROC scores for the Kuramoto dataset, AUROC scores for the highest scoring CPNNs on the River benchmarks, and Accuracy for the Aerosol-Cloud benchmark. The highest scores are denoted in \textbf{bold}. We additionally report the variance, calculated over 10 runs on the Kuramoto dataset.in brackets. * \cite{lowe_amortized_2022}.}
\label{tab:3}
\end{table*}

    

    

        

        

\begin{figure}[{h}]
\includegraphics[width=0.98\linewidth]{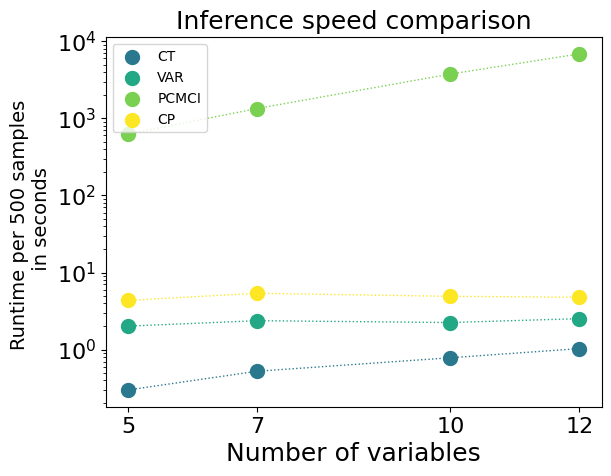}
\caption{Comparison of inference speed. Here, we use the architecture with the highest AUROC score for each dataset to represent CPNNs. We report the speed of computing the solution for 500 samples and 100 repetitions.}
\label{g:5}
\end{figure}

As we investigated the relationship between the training distribution and the performance of CP, we also wanted to explore the relationship between the number of parameters and the ability to learn increasingly more complex datasets. For this, we generated three synthetic linear datasets and stepwise increased the number of variables to 10. We performed Causal Pretraining on each dataset, trained varying model sizes independently, and performed the exact hyperparameter search as in the previous chapter. We selected the performance of the best-scoring hyperparameter combination of the best-performing architecture. We then retrained this configuration five times to calculate the final reference scores reported in \autoref{g:5}. Note that for the size "lcm" we restricted our search to the Trf and the CM architecture. 
We find that the parameter count of the model determines the ability to perform causal discovery properly. Specifically, there seems to be a parameter threshold under which no proper solution can be learned (AUROC 0.5). We also find that the extrapolation capabilities increase even when keeping the same number of data samples (contrary to what one would typically expect concerning overfitting). Further, the required model size for certain variables was much bigger than expected. However, we suggest that this might be partly because discovering a causal graph $G$ with multiple lags requires the estimation of $V \times V \times N $ values, which might be hard for CPNNs. This suggests that reducing the number of lags might reduce the required parameters in the future. Together, we conclude from this that to improve the performance of CP, we need to increase its scaling, especially considering that it is likely that learning much broader distributions will also require an even higher parameter count.

 \subsection{Kuramoto}
 To bridge the gap to \cite{lowe_amortized_2022}, we evaluate CP on data from a Kuramoto Model with five variables. We generated 50,000 samples for training and 5,000 samples for testing, originating from a simulation of this model. Further, we kept the same simulation settings as in \cite{lowe_amortized_2022} and ignored the diagonal of $G$ (autoregressive links) to make our experiments directly comparable. We again performed a hyperparameter search for each architecture and retrained the best hyperparameter combination ten times to report the standard deviation. We report the results of these experiments in \autoref{tab:3}.
 Here, the \textbf{CM} architectures performed best and achieved similar, if not slightly better, results than \cite{lowe_amortized_2022}. Further, all of our baselines are decisively outperformed on this dataset, which we attribute to the fact that the Kuramoto dataset does not follow \autoref{eq:1}, which hinders methods that rely on linear relationships.

\subsection{Zero-shot Inference}

To test the performance of CPNNs when applied to truly out-of-distribution data, we reuse the CDNNs from  \hyperref[synth:1]{Synthetic Data 1~\ref*{synth:1}} on two benchmarks involving the discharge of rivers in Germany \cite{ahmad_causal_2022} 
and aerosol-cloud interactions \cite{jesson_using_2021}. Since all of these datasets only specify a summary graph $G$, we took the causal link of the first time lag as the prediction for the summary graph. We report the results of this experiment in \autoref{tab:3}. Importantly, since we wanted to perform this experiment as close to a real application of CPNNs as possible, we do not report STD but inferred the summary graph once for each sample from the full-time series and with the highest-scoring models from  \hyperref[synth:1]{Synthetic Data 1~\ref*{synth:1}}.

We found that the performance of CPNNs on unseen data distribution differs widely between model architectures and datasets. While the river benchmark is properly predicted by the \textbf{uGRU} and the \textbf{Trf} architecture, these architectures are specifically outperformed by \textbf{CM} on the aerosol dataset. 
As a side note, \cite{ahmad_causal_2022} or originally \cite{gerhardus_high-recall_2020} suggest that the river benchmark is more challenging than our scoring suggests. This is, since they report results for fixed p-values, which is not required when calculating AUROC scores.
While the performance of CP is improvable, we find it promising that models not trained on these datasets do not break down entirely but display some extrapolation capabilities. We believe this signifies that CPNNs incorporate some form of Causal Discovery test that can uncover correct causal structures in more general settings than the training distribution. 

\section{Discussion and Conclusion}

This work introduced Causal Pretraining, a methodology for supervised deep end-to-end causal discovery from time series data.
We conducted the first experiments, providing evidence that our causally pre-trained neural networks (CDNNs) can achieve similar results to other causal discovery methods without fine-tuning any parameters during inference and being highly computationally parallelizable.
We provided evidence that CDNNs show some potential to extrapolate to unseen real-life data outside of the training distribution.
Specifically interesting, the performance of CDNNs and its capability to generalize increases with data complexity and model size. Under the assumption that this trend continues beyond the scope of our experiments, we take this as evidence that foundation causal models trained through Causal Pretraining are possible. Recent developments in deep learning \cite{openai_gpt-4_2023} suggest that consistent generalization by scaling up through data and model dimensions is indeed possible. This idea also aligns with theoretical research such as \cite{bubeck_universal_2021}, \cite{bartlett_benign_2020} or \cite{brutzkus_why_2019}, suggesting that greatly over-parametrizing neural networks can help with generalization. While we excluded these experiments in this paper since they are probably intuitive for deep-learning practitioners, we also found that both scalings must go hand-in-hand while performing Causal Pretraining. Simply scaling up the architectures without increasing the dataset's complexity makes it arbitrarily easy for neural networks to remember the training dataset and not generalize at all.
Further, keeping the network too small while increasing the data complexity in might make the learning task impossible and, with that, also prevent generalization. We reserve keeping this delicate balance while scaling up our methodology concerning data distribution and model size for future research. 



\newpage

\section{Acknowledgments}
We gratefully recognize the support of iDiv\textsuperscript{1}, which is funded
by the German Research Foundation (DFG – FZT 118, 202548816).
Special thanks from Gideon Stein to Yuanyuan Huang\textsuperscript{1}, Anne Ebeling\textsuperscript{1}, and Nico Eisenhauer\textsuperscript{1} for supporting me in this project and giving me the freedom to be creative with my research ideas. Special thanks also to Christian Reimers \textsuperscript{2} for the creative and helpful exchange of ideas.

Gideon Stein is funded by the iDiv flexpool (No 06203674-22).
Maha Shadaydeh is  funded by the Carl Zeiss Foundation within the scope of the program line “Breakthroughs: Exploring Intelligent Systems” for “Digitization—explore the basics (No P2017-01-003), use applications”.
\begin{center}
    \textit{\textsuperscript{\rm 1}German Centre of Integrative Biodiversity Research} \linebreak
    \textit{\textsuperscript{\rm 2}Max Planck Institute for Biogeochemistry Jena}
\end{center}

 \bibliography{aaai24}

\begin{thebibliography}{46}
\providecommand{\natexlab}[1]{#1}

\bibitem[{Ahmad, Shadaydeh, and Denzler(2022)}]{ahmad_causal_2022}
Ahmad, W.; Shadaydeh, M.; and Denzler, J. 2022.
\newblock Causal {Discovery} using {Model} {Invariance} through {Knockoff} {Interventions}.
\newblock In \emph{{ICML} 2022: {Workshop} on {Spurious} {Correlations}, {Invariance} and {Stability}}.

\bibitem[{Akaike(1992)}]{kotz_information_1992}
Akaike, H. 1992.
\newblock Information {Theory} and an {Extension} of the {Maximum} {Likelihood} {Principle}.
\newblock 610--624. New York, NY: Springer New York.
\newblock ISBN 9780387940373 9781461209195.

\bibitem[{Amodei et~al.(2016)Amodei, Ananthanarayanan, Anubhai, Bai, Battenberg, Case, Casper, Catanzaro, Cheng, Chen, Chen, Chen, Chen, Chrzanowski, Coates, Diamos, Ding, Du, Elsen, Engel, Fang, Fan, Fougner, Gao, Gong, Hannun, Han, Johannes, Jiang, Ju, Jun, LeGresley, Lin, Liu, Liu, Li, Li, Ma, Narang, Ng, Ozair, Peng, Prenger, Qian, Quan, Raiman, Rao, Satheesh, Seetapun, Sengupta, Srinet, Sriram, Tang, Tang, Wang, Wang, Wang, Wang, Wang, Wang, Wu, Wei, Xiao, Xie, Xie, Yogatama, Yuan, Zhan, and Zhu}]{amodei_deep_2016}
Amodei, D.; Ananthanarayanan, S.; Anubhai, R.; Bai, J.; Battenberg, E.; Case, C.; Casper, J.; Catanzaro, B.; Cheng, Q.; Chen, G.; Chen, J.; Chen, J.; Chen, Z.; Chrzanowski, M.; Coates, A.; Diamos, G.; Ding, K.; Du, N.; Elsen, E.; Engel, J.; Fang, W.; Fan, L.; Fougner, C.; Gao, L.; Gong, C.; Hannun, A.; Han, T.; Johannes, L.; Jiang, B.; Ju, C.; Jun, B.; LeGresley, P.; Lin, L.; Liu, J.; Liu, Y.; Li, W.; Li, X.; Ma, D.; Narang, S.; Ng, A.; Ozair, S.; Peng, Y.; Prenger, R.; Qian, S.; Quan, Z.; Raiman, J.; Rao, V.; Satheesh, S.; Seetapun, D.; Sengupta, S.; Srinet, K.; Sriram, A.; Tang, H.; Tang, L.; Wang, C.; Wang, J.; Wang, K.; Wang, Y.; Wang, Z.; Wang, Z.; Wu, S.; Wei, L.; Xiao, B.; Xie, W.; Xie, Y.; Yogatama, D.; Yuan, B.; Zhan, J.; and Zhu, Z. 2016.
\newblock Deep {Speech} 2 : {End}-to-{End} {Speech} {Recognition} in {English} and {Mandarin}.
\newblock In \emph{Proceedings of {The} 33rd {International} {Conference} on {Machine} {Learning}}, 173--182. PMLR.

\bibitem[{Assaad, Devijver, and Gaussier(2022)}]{assaad_survey_2022}
Assaad, C.~K.; Devijver, E.; and Gaussier, E. 2022.
\newblock Survey and {Evaluation} of {Causal} {Discovery} {Methods} for {Time} {Series}.
\newblock \emph{Journal of Artificial Intelligence Research}, 73: 767--819.

\bibitem[{Bartlett et~al.(2020)Bartlett, Long, Lugosi, and Tsigler}]{bartlett_benign_2020}
Bartlett, P.~L.; Long, P.~M.; Lugosi, G.; and Tsigler, A. 2020.
\newblock Benign {Overfitting} in {Linear} {Regression}.
\newblock \emph{Proceedings of the National Academy of Sciences}, 117(48): 30063--30070.
\newblock ArXiv:1906.11300 [cs, math, stat].

\bibitem[{Bojarski et~al.(2016)Bojarski, Del~Testa, Dworakowski, Firner, Flepp, Goyal, Jackel, Monfort, Muller, Zhang, Zhang, Zhao, and Zieba}]{bojarski_end_2016}
Bojarski, M.; Del~Testa, D.; Dworakowski, D.; Firner, B.; Flepp, B.; Goyal, P.; Jackel, L.~D.; Monfort, M.; Muller, U.; Zhang, J.; Zhang, X.; Zhao, J.; and Zieba, K. 2016.
\newblock End to {End} {Learning} for {Self}-{Driving} {Cars}.
\newblock ADS Bibcode: 2016arXiv160407316B.

\bibitem[{Brutzkus and Globerson(2019)}]{brutzkus_why_2019}
Brutzkus, A.; and Globerson, A. 2019.
\newblock Why do {Larger} {Models} {Generalize} {Better}? {A} {Theoretical} {Perspective} via the {XOR} {Problem}.
\newblock In \emph{Proceedings of the 36th {International} {Conference} on {Machine} {Learning}}, 822--830. PMLR.

\bibitem[{Bubeck and Sellke(2021)}]{bubeck_universal_2021}
Bubeck, S.; and Sellke, M. 2021.
\newblock A {Universal} {Law} of {Robustness} via {Isoperimetry}.
\newblock In \emph{Advances in {Neural} {Information} {Processing} {Systems}}, volume~34, 28811--28822. Curran Associates, Inc.

\bibitem[{Dang, Shah, and Zerfos(2018)}]{dang_seq2graph_2018}
Dang, X.-H.; Shah, S.~Y.; and Zerfos, P. 2018.
\newblock seq2graph: {Discovering} {Dynamic} {Dependencies} from {Multivariate} {Time} {Series} with {Multi}-level {Attention}.
\newblock ArXiv:1812.04448 [cs, stat].

\bibitem[{Geffner et~al.(2022)Geffner, Antoran, Foster, Gong, Ma, Kiciman, Sharma, Lamb, Kukla, Hilmkil, Jennings, Pawlowski, Allamanis, and Zhang}]{geffner_deep_2022}
Geffner, T.; Antoran, J.; Foster, A.; Gong, W.; Ma, C.; Kiciman, E.; Sharma, A.; Lamb, A.; Kukla, M.; Hilmkil, A.; Jennings, J.; Pawlowski, N.; Allamanis, M.; and Zhang, C. 2022.
\newblock Deep {End}-to-end {Causal} {Inference}.

\bibitem[{Gerhardus and Runge(2020)}]{gerhardus_high-recall_2020}
Gerhardus, A.; and Runge, J. 2020.
\newblock High-recall causal discovery for autocorrelated time series with latent confounders.
\newblock In \emph{Advances in {Neural} {Information} {Processing} {Systems}}, volume~33, 12615--12625. Curran Associates, Inc.

\bibitem[{Glasmachers(2017)}]{glasmachers_limits_2017}
Glasmachers, T. 2017.
\newblock Limits of {End}-to-{End} {Learning}.
\newblock In \emph{Proceedings of the {Ninth} {Asian} {Conference} on {Machine} {Learning}}, 17--32. PMLR.

\bibitem[{Gong et~al.(2023)Gong, Yao, Zhang, Li, and Bi}]{gong_causal_2023}
Gong, C.; Yao, D.; Zhang, C.; Li, W.; and Bi, J. 2023.
\newblock Causal {Discovery} from {Temporal} {Data}: {An} {Overview} and {New} {Perspectives}.
\newblock ArXiv:2303.10112 [cs, stat].

\bibitem[{Granger(1969)}]{granger_investigating_1969}
Granger, C. W.~J. 1969.
\newblock Investigating {Causal} {Relations} by {Econometric} {Models} and {Cross}-spectral {Methods}.
\newblock \emph{Econometrica}, 37(3): 424--438.

\bibitem[{Guo, Lin, and Antulov-Fantulin(2019)}]{guo_exploring_2019}
Guo, T.; Lin, T.; and Antulov-Fantulin, N. 2019.
\newblock Exploring interpretable {LSTM} neural networks over multi-variable data.
\newblock In \emph{Proceedings of the 36th {International} {Conference} on {Machine} {Learning}}, 2494--2504. PMLR.

\bibitem[{Hornik, Stinchcombe, and White(1989)}]{hornik_multilayer_1989}
Hornik, K.; Stinchcombe, M.; and White, H. 1989.
\newblock Multilayer feedforward networks are universal approximators.
\newblock \emph{Neural Networks}, 2(5): 359--366.

\bibitem[{Huang et~al.(2018)Huang, Krueger, Lacoste, and Courville}]{huang_neural_2018}
Huang, C.-W.; Krueger, D.; Lacoste, A.; and Courville, A. 2018.
\newblock Neural {Autoregressive} {Flows}.
\newblock In \emph{Proceedings of the 35th {International} {Conference} on {Machine} {Learning}}, 2078--2087. PMLR.

\bibitem[{Jesson et~al.(2021)Jesson, Manshausen, Douglas, Watson-Parris, Gal, and Stier}]{jesson_using_2021}
Jesson, A.; Manshausen, P.; Douglas, A.; Watson-Parris, D.; Gal, Y.; and Stier, P. 2021.
\newblock Using {Non}-{Linear} {Causal} {Models} to {Study} {Aerosol}-{Cloud} {Interactions} in the {Southeast} {Pacific}.
\newblock ArXiv:2110.15084 [physics].

\bibitem[{Kuramoto(1975)}]{kuramoto_self-entrainment_1975}
Kuramoto, Y. 1975.
\newblock Self-entrainment of a population of coupled non-linear oscillators.
\newblock \emph{Mathematical Problems in Theoretical Physics}, 39: 420--422.
\newblock ADS Bibcode: 1975LNP....39..420K.

\bibitem[{Kyono, Zhang, and van~der Schaar(2020)}]{kyono_castle_2020}
Kyono, T.; Zhang, Y.; and van~der Schaar, M. 2020.
\newblock {CASTLE}: {Regularization} via {Auxiliary} {Causal} {Graph} {Discovery}.
\newblock In \emph{Advances in {Neural} {Information} {Processing} {Systems}}, volume~33, 1501--1512. Curran Associates, Inc.

\bibitem[{Lachapelle et~al.(2020)Lachapelle, Brouillard, Deleu, and Lacoste-Julien}]{lachapelle_gradient-based_2020}
Lachapelle, S.; Brouillard, P.; Deleu, T.; and Lacoste-Julien, S. 2020.
\newblock Gradient-{Based} {Neural} {DAG} {Learning}.
\newblock ArXiv:1906.02226 [cs, stat].

\bibitem[{Li, Xiao, and Tian(2020)}]{li_supervised_2020}
Li, H.; Xiao, Q.; and Tian, J. 2020.
\newblock Supervised {Whole} {DAG} {Causal} {Discovery}.
\newblock ArXiv:2006.04697 [cs, stat].

\bibitem[{Li, Yu, and Principe(2023)}]{li_causal_2023}
Li, H.; Yu, S.; and Principe, J. 2023.
\newblock Causal {Recurrent} {Variational} {Autoencoder} for {Medical} {Time} {Series} {Generation}.
\newblock ArXiv:2301.06574 [cs, eess].

\bibitem[{Lopez-Paz et~al.(2015)Lopez-Paz, Muandet, Schölkopf, and Tolstikhin}]{lopez-paz_towards_2015}
Lopez-Paz, D.; Muandet, K.; Schölkopf, B.; and Tolstikhin, I. 2015.
\newblock Towards a {Learning} {Theory} of {Cause}-{Effect} {Inference}.
\newblock In \emph{Proceedings of the 32nd {International} {Conference} on {Machine} {Learning}}, 1452--1461. PMLR.

\bibitem[{Loshchilov and Hutter(2018)}]{loshchilov_decoupled_2018}
Loshchilov, I.; and Hutter, F. 2018.
\newblock Decoupled {Weight} {Decay} {Regularization}.

\bibitem[{Löwe et~al.(2022)Löwe, Madras, Zemel, and Welling}]{lowe_amortized_2022}
Löwe, S.; Madras, D.; Zemel, R.; and Welling, M. 2022.
\newblock Amortized {Causal} {Discovery}: {Learning} to {Infer} {Causal} {Graphs} from {Time}-{Series} {Data}.
\newblock In \emph{Proceedings of the {First} {Conference} on {Causal} {Learning} and {Reasoning}}, 509--525. PMLR.

\bibitem[{Meng(2019)}]{meng_estimating_2019}
Meng, Y. 2019.
\newblock Estimating {Granger} {Causality} with {Unobserved} {Confounders} via {Deep} {Latent}-{Variable} {Recurrent} {Neural} {Network}.
\newblock ArXiv:1909.03704 [cs, stat].

\bibitem[{Montalto et~al.(2015)Montalto, Stramaglia, Faes, Tessitore, Prevete, and Marinazzo}]{montalto_neural_2015}
Montalto, A.; Stramaglia, S.; Faes, L.; Tessitore, G.; Prevete, R.; and Marinazzo, D. 2015.
\newblock Neural {Networks} with {Non}-{Uniform} {Embedding} and {Explicit} {Validation} {Phase} to {Assess} {Granger} {Causality}.
\newblock \emph{Neural Networks}, 71: 159--171.

\bibitem[{Ng et~al.(2019)Ng, Zhu, Chen, and Fang}]{ng_graph_2019}
Ng, I.; Zhu, S.; Chen, Z.; and Fang, Z. 2019.
\newblock A {Graph} {Autoencoder} {Approach} to {Causal} {Structure} {Learning}.
\newblock ADS Bibcode: 2019arXiv191107420N.

\bibitem[{{OpenAI}(2023)}]{openai_gpt-4_2023}
{OpenAI}. 2023.
\newblock {GPT}-4 {Technical} {Report}.
\newblock ArXiv:2303.08774 [cs].

\bibitem[{Pamfil et~al.(2020)Pamfil, Sriwattanaworachai, Desai, Pilgerstorfer, Georgatzis, Beaumont, and Aragam}]{pamfil_dynotears_2020}
Pamfil, R.; Sriwattanaworachai, N.; Desai, S.; Pilgerstorfer, P.; Georgatzis, K.; Beaumont, P.; and Aragam, B. 2020.
\newblock {DYNOTEARS}: {Structure} {Learning} from {Time}-{Series} {Data}.
\newblock In \emph{Proceedings of the {Twenty} {Third} {International} {Conference} on {Artificial} {Intelligence} and {Statistics}}, 1595--1605. PMLR.

\bibitem[{Petersen et~al.(2023)Petersen, Ramsey, Ekstrøm, and Spirtes}]{petersen_causal_2023}
Petersen, A.~H.; Ramsey, J.; Ekstrøm, C.~T.; and Spirtes, P. 2023.
\newblock Causal {Discovery} for {Observational} {Sciences} {Using} {Supervised} {Machine} {Learning}.
\newblock \emph{Journal of Data Science}, 21(2): 255--280.

\bibitem[{Runge et~al.(2023)Runge, Gerhardus, Varando, Eyring, and Camps-Valls}]{runge_causal_2023}
Runge, J.; Gerhardus, A.; Varando, G.; Eyring, V.; and Camps-Valls, G. 2023.
\newblock Causal inference for time series.
\newblock \emph{Nature Reviews Earth \& Environment}, 4(7): 487--505.

\bibitem[{Runge et~al.(2019)Runge, Nowack, Kretschmer, Flaxman, and Sejdinovic}]{runge_detecting_2019}
Runge, J.; Nowack, P.; Kretschmer, M.; Flaxman, S.; and Sejdinovic, D. 2019.
\newblock Detecting causal associations in large nonlinear time series datasets.
\newblock \emph{Science Advances}, 5(11): eaau4996.
\newblock ArXiv:1702.07007 [physics, stat].

\bibitem[{Shah and Peters(2020)}]{shah_hardness_2020}
Shah, R.~D.; and Peters, J. 2020.
\newblock The {Hardness} of {Conditional} {Independence} {Testing} and the {Generalised} {Covariance} {Measure}.
\newblock \emph{The Annals of Statistics}, 48(3).
\newblock ArXiv:1804.07203 [math, stat].

\bibitem[{Spirtes et~al.(2000)Spirtes, Glymour, Scheines, and Heckerman}]{spirtes_causation_2000}
Spirtes, P.; Glymour, C.~N.; Scheines, R.; and Heckerman, D. 2000.
\newblock \emph{Causation, {Prediction}, and {Search}}.
\newblock MIT Press.
\newblock ISBN 9780262194402.

\bibitem[{Sun et~al.(2023)Sun, Schulte, Liu, and Poupart}]{sun_nts-notears_2023}
Sun, X.; Schulte, O.; Liu, G.; and Poupart, P. 2023.
\newblock {NTS}-{NOTEARS}: {Learning} {Nonparametric} {DBNs} {With} {Prior} {Knowledge}.
\newblock In \emph{Proceedings of {The} 26th {International} {Conference} on {Artificial} {Intelligence} and {Statistics}}, 1942--1964. PMLR.

\bibitem[{Tank et~al.(2018)Tank, Cover, Foti, Shojaie, and Fox}]{tank_interpretable_2018}
Tank, A.; Cover, I.; Foti, N.~J.; Shojaie, A.; and Fox, E.~B. 2018.
\newblock An {Interpretable} and {Sparse} {Neural} {Network} {Model} for {Nonlinear} {Granger} {Causality} {Discovery}.
\newblock ArXiv:1711.08160 [stat].

\bibitem[{Tank et~al.(2021)Tank, Covert, Foti, Shojaie, and Fox}]{tank_neural_2021}
Tank, A.; Covert, I.; Foti, N.; Shojaie, A.; and Fox, E. 2021.
\newblock Neural {Granger} {Causality}.
\newblock \emph{IEEE Transactions on Pattern Analysis and Machine Intelligence}, 1--1.
\newblock ArXiv:1802.05842 [stat].

\bibitem[{Taud and Mas(2018)}]{taud_multilayer_2018}
Taud, H.; and Mas, J. 2018.
\newblock Multilayer {Perceptron} ({MLP}).
\newblock In Camacho~Olmedo, M.~T.; Paegelow, M.; Mas, J.-F.; and Escobar, F., eds., \emph{Geomatic {Approaches} for {Modeling} {Land} {Change} {Scenarios}}, Lecture {Notes} in {Geoinformation} and {Cartography}, 451--455. Cham: Springer International Publishing.
\newblock ISBN 9783319608013.

\bibitem[{Teodora~Trifunov, Shadaydeh, and Denzler(2022)}]{teodora_trifunov_time_2022}
Teodora~Trifunov, V.; Shadaydeh, M.; and Denzler, J. 2022.
\newblock Time {Series} {Causal} {Link} {Estimation} under {Hidden} {Confounding} using {Knockoff} {Interventions}.
\newblock ADS Bibcode: 2022arXiv220911497T.

\bibitem[{Trockman and Zico~Kolter(2022)}]{trockman_patches_2022}
Trockman, A.; and Zico~Kolter, J. 2022.
\newblock Patches {Are} {All} {You} {Need}?
\newblock ADS Bibcode: 2022arXiv220109792T.

\bibitem[{Vowels, Camgoz, and Bowden(2022)}]{vowels_dx2019ya_2022}
Vowels, M.~J.; Camgoz, N.~C.; and Bowden, R. 2022.
\newblock D\&\#x2019;ya {Like} {DAGs}? {A} {Survey} on {Structure} {Learning} and {Causal} {Discovery}.
\newblock \emph{ACM Computing Surveys}, 55(4): 82:1--82:36.

\bibitem[{Wang et~al.(2018)Wang, Lin, Qi, Lian, Feng, Wu, and Pan}]{wang_estimating_2018}
Wang, Y.; Lin, K.; Qi, Y.; Lian, Q.; Feng, S.; Wu, Z.; and Pan, G. 2018.
\newblock Estimating {Brain} {Connectivity} {With} {Varying}-{Length} {Time} {Lags} {Using} a {Recurrent} {Neural} {Network}.
\newblock \emph{IEEE Transactions on Biomedical Engineering}, 65(9): 1953--1963.

\bibitem[{Zheng et~al.(2018)Zheng, Aragam, Ravikumar, and Xing}]{zheng_dags_2018}
Zheng, X.; Aragam, B.; Ravikumar, P.; and Xing, E. 2018.
\newblock {DAGs} with {NO} {TEARS}: {Smooth} {Optimization} for {Structure} {Learning}.

\bibitem[{Zhou et~al.(2021)Zhou, Zhang, Peng, Zhang, Li, Xiong, and Zhang}]{zhou_informer_2021}
Zhou, H.; Zhang, S.; Peng, J.; Zhang, S.; Li, J.; Xiong, H.; and Zhang, W. 2021.
\newblock Informer: {Beyond} {Efficient} {Transformer} for {Long} {Sequence} {Time}-{Series} {Forecasting}.
\newblock \emph{Proceedings of the AAAI Conference on Artificial Intelligence}, 35(12): 11106--11115.
\newblock Number: 12.

\end{thebibliography}

\clearpage
\section{Appendix}
\label{sec:app}
Here, we include additional specifications for the exact parameter counts for all used architectures and sizes in \autoref{app:3} and data that we use throughout this paper in   \hyperref[app:1]{Algorithm~\ref*{app:1}} and \autoref{app:2}. Next, we include a study on inference speed in \hyperref[ablation:0]{Inference Speed~\ref*{ablation:0}} and a depiction of Correlation Regularization in \autoref{g:3}. While we expected that the importance of hyperparameter selection would decrease with model size, we found no consistent pattern in our experiments. To depict this, we display an example of performance distribution over hyperparameter in \autoref{app:7}. Further, in \hyperref[ablation:1]{Preliminary studies~\ref*{ablation:1}}, we describe two small experiments on the baseline capabilities of our architectures, which were historically performed before experiments in the \hyperref[experiments:1]{Experiment section~\ref*{experiments:1}}. In \hyperref[ablation:2]{Probabilistic studies~\ref*{ablation:2}}, we introduce a procedure that allows CPNNs to predict probability distributions instead of probability scalars that might be useful for certain applications. Finally, we provide a summary of all abbreviations used throughout the paper in \autoref{app:8}.

\begin{algorithm}
\caption{An algorithm with caption}\label{alg:cap}
\label{app:1}

\begin{algorithmic}

\Require: $maximum\char`_lags(n), number\char`_of\char`_variables(v)$ 
\Require: $link\char`_threshold, link\char`_distribution$ 
\Require: $ts\char`_length,  noise\char`_var, number\char`_of\char`_samples$ 
\Require: $nl\_selection$, $nl\_threshold$
\Require: $data\char`_test(), graph\char`_test()$ 

\State $\mathcal{X} = \{ \}$
\State $\mathcal{G} = \{ \}$

\While{$len(\mathcal{G}) < number\char`_of\char`_samples$}
    \State $Draw: U(0,1)^{V \times V \times N}$
    \State $PA:  1 \; \textbf{if} \; Draw > link\char`_threshold\; \textbf{else}\; 0$
    \State $\mathbb{A} = U(link\_distribution)\; \textbf{if} \;PA\;\textbf{else}\; 0$
    \If{nonlinear} 
        \State $Draw:   U(0,1)^{V \times V \times N}$
        \State $PA:  1 \; \textbf{if} \; Draw > nl\_threshold\; \textbf{else}\; 0$
        \State $f: U(nl\_selection) \; \textbf{if}  \;PA  \;\textbf{else} \; 1$
    \Else
        \State $f: 1$
    \EndIf
        \State $\mathbb{X} = \{\}$
        \While{$len(\mathbb{X}) < ts\_length$}
            \State generate $x^{new}_i$ according to \autoref{eq:1} $(\mathbb{A}, f)$
        \EndWhile
        \If{$data\char`_test(\mathbb{X})$}
            \State $\mathcal{X} \leftarrow X$ 
            \State $\mathcal{G} \leftarrow G$
        \EndIf
\EndWhile

\end{algorithmic}
\end{algorithm}

\begin{table}[h]
    \centering
    \begin{tabular}{lccccc}
    \toprule
        Size &\textbf{ Small} & \textbf{Medium} & \textbf{Big} & \textbf{Deep} & \textbf{LCM} \\ 
    \midrule
        \textbf{MLP} & 13.4k & 118k & 1.5M & 17.3M & -\\
       \textbf{uGRU}& 13.0k & 126k & 1.6M & 17.2M &-\\ 
        \textbf{bGRU} & 12.7k & 116k & 1.5M & 18.4M & -  \\ 
        \textbf{CM} & 13.3k & 128k & 1.4M &  17M & 286M\\ 
        \textbf{Trf} & 12.3k & 120k & 1.4M & 16.1M & 391M\\
    \bottomrule
    \end{tabular}
    \caption{Excact parameter count for all architectures and sizes we use throughout this paper.}
\label{app:3}
\end{table}

\subsection{Inference Speed}
\label{ablation:0}

Since inference with CPNNs is highly parallelizable (batching), which we denote as a neat advantage of CPNNs, we provide some short comparisons on inference speed.
For this, we reuse the dataset of the above section and measure the speed to predict the corresponding test set with 500 samples each. We perform inference on an Intel i7-7700 CPU, omitting GPU support from which CPNNs would additionally benefit. We summarize the relationship between the number of variables and the inference speed as follows:
While our simple benchmark techniques are still slightly faster since they come with closed-form solutions for their optimization (and we additionally implemented a batched version for \textbf{CT}), the inference speed of Causal Pretraining lies in the same order of magnitude for all number of variables tested.
In contrast, \textbf{PCMCI} scales quadratically in our experiments. The number of conditional independence tests that must be performed grows quadratically with the number of variables, making it infeasible to infer from a large set of variables. Importantly, all our methods (including \textbf{PCMCI}) are very fast to compute in comparison to other approaches such as, e.g., \cite{tank_interpretable_2018} or \cite{ahmad_causal_2022}, which require fitting neural networks for each sample. We believe this feature of CPNNs makes it specifically beneficial in areas where many similar samples have to be processed, such as in Earth and neuroscience. Especially considering that they can outperform similarly fast benchmarks.

\begin{table*}
    \centering
    \begin{tabular}{lcccccccc}
    \toprule
    
        \textbf{Name} & \textbf{Pre} & \textbf{SL} & \textbf{ML} & \textbf{SNL} & \textbf{MNL} & \textbf{LNL} & \textbf{XLNL} & \textbf{Wide} \\ 
        \midrule
        Function set  & L & L & L  & NL1 & NL1 & NL2 & NL2 & NL2\\
        Variables  & 3 & 3 & 5 & 3 & 5 & 3 & 5 & 7/10/15\\ 
        Maximum lag  & 2 & 2 & 3 & 2 & 3 & 2 & 3 & 3 \\
        Coefficients (train)& br &  br & $\pm$ br &  br & $\pm$ br &  br & $\pm$ br & $\pm$ br\\
        Noise var (train) & 0.4 & 0.4 & 0.4 & 0.4 & 0.4 & 0.4 & 0.4 & 0.4 \\ 
        Coefficients (test 1)  & br &  br &  $\pm$ br & br &  $\pm$ br &  br &$\pm$  br & $\pm$ br \\ 
        Noise var (test 1 & 0.4 & 0.4 & 0.4 & 0.4 & 0.4 & 0.4 & 0.4 & 0.4 \\
        Coefficients (test 2)  & sr & sr &  $\pm$ sr& sr &  $\pm$ sr & sr &  $\pm$ sr &  $\pm$ sr \\ 
        Noise var (test 2)  & 0.6 & 0.6 & 0.6 & 0.6 & 0.6 & 0.6 & 0.6  & 0.6 \\
        Average non-zero & 9 & 2.7 & 7.5 & 2.7 & 7.5 & 2.7 & 7.5 & 14.7/30/67.5\\ \bottomrule
    \end{tabular}
    \caption{Dataset specifications. L denotes linear, NL1 denotes drawing only from ($e^x$, $x^2$), and NL2 denotes the full function set specified in the method \hyperref[sec:method]{Method section~\ref*{sec:method}}. Furthermore, sr denotes the range (0.2-0.3), and br denotes the range (0.3-0.5). $\pm$ denotes that values drawn from a range can be positive or negative. \textbf{Wide} specifies the datasets that were used during  \hyperref[synth:2]{Synthetic Data 2~\ref*{synth:1}}. \textbf{Pre} specifies the dataset used during \hyperref[ablation:1]{Preliminary studies~\ref*{ablation:1}}.}
\label{app:2}
\end{table*}

\subsection{Preliminary Studies: Architecture Capabilities}
\label{ablation:1}
Initially, we conducted two experiments to estimate our architecture's general capabilities. We used the model size "small" for these three experiments and chose a learning rate of 0.0001, a weight decay of 0.01, and a batch size of 64, performing no hyperparameter search.

First, we tested whether our architectures can correctly distinguish between a linear and a nonlinear $\mathbb{X}$, which we deemed to be an essential capability.
For this purpose, we fixed which elements of $\mathbb{A}$ are non-zero and only alter their actual values. We then added a nonlinear $f^n_{j, i}$ for half of the samples as described in \hyperref[app:1]{Algorithm~\ref*{app:1}}.
We temporarily re-purposed the regression head by adding a Sigmoid function.
We treated the output as a binary classification and used binary cross entropy as the loss function.

We found that the \textbf{uGRU} and the \textbf{Trf} architecture outperform the other approaches in this limited setting (Increased AUROC scores on Test data), framing them as the most promising candidates. While we found this to be rather intuitive, since these are archetypes frequently applied to time series, we also later found out that this advantage does not necessarily translate into more general setups. We conclude that this supports our approach of considering multiple architectures.

Second, we tested whether CPNNs can uncover the precise values in $\mathbb{A}$. We again fix which values of $\mathbb{A}$ are non-zero and alter only its coefficients.
To train, we remove the sigmoid activation from our output vector and treat it as a direct prediction for $\mathbb{A}$.
We take the original $\mathbb{A}$ as labels and optimize an MSE loss.

We found that none of our architectures can uncover the exact $\mathbb{A}$. They instead converge on predicting the mean of each element in $\mathbb{A}$. This behavior is consistent over all tested architectures, and we observed no change even when weighing the loss values of non-zero elements higher than those elements that are zero. We took this as a reason for our binary problem formulation, aiming only at uncovering which elements are non-zero but not the exact strength of the relationship. While we are confident that this behavior might be altered with more sophisticated loss schemes or larger model sizes, we keep this additional complexity for future research. Importantly, making binary decisions aligns with many other causal inference methods.



\subsection{Probabilistic Predictions}
\label{ablation:2}
\begin{figure*}
\centering
\includegraphics[width=0.99\linewidth]{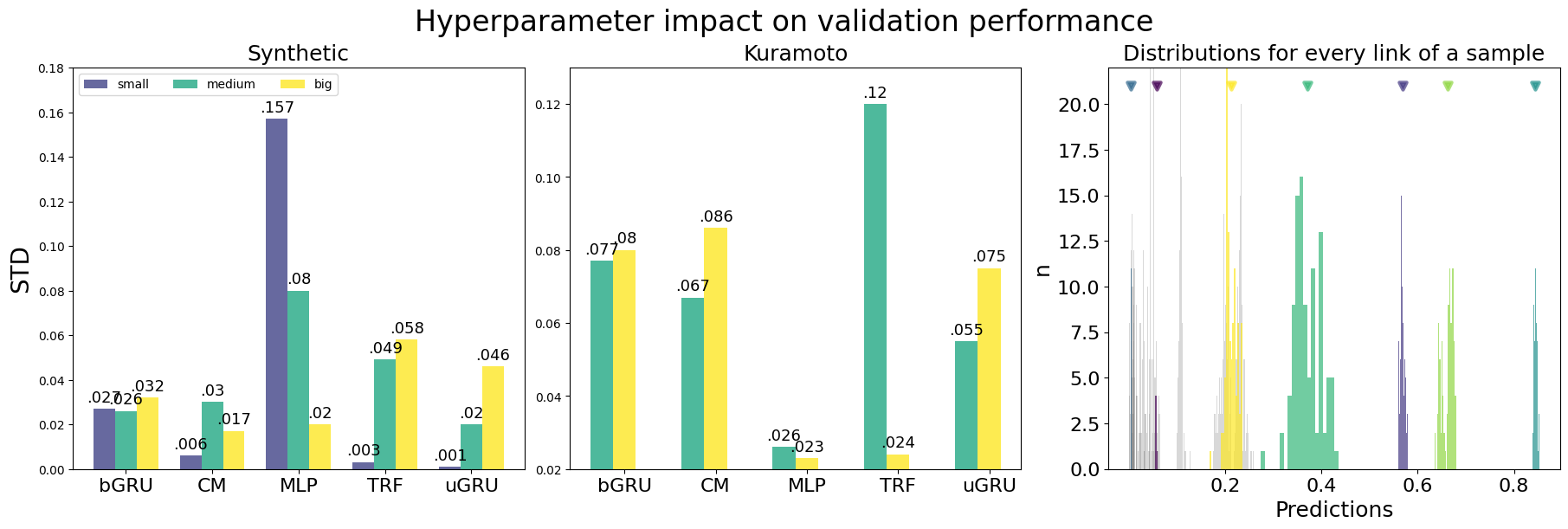}
\caption{\textbf{Left:} Impact of hyperparameter selection. We report the standard deviation of the AUROC score on the validation set. For the synthetic data, we provide the joint dataset grid search results for model sizes "medium" and "big" and the grid search results for the dataset \textbf{LnL} for model size "small," ignoring all combinations that were not evaluated for the joint dataset. \textbf{Right:} Distributions over output probabilities for every edge in a single sample, sampled as described in \hyperref[ablation:2]{Probabilistic studies~\ref*{ablation:2}}. Colorful distributions denote true edges.}
\label{app:7}
\end{figure*}

Here, we tested whether we can adapt our method to produce distribution as outputs instead of a single probability. We do this by running many short samples of the same time series through a model and constructing a distribution over output probabilities for each edge. It might be interesting to analyze the form of these distributions in the future, which could help detect wrongly classified edges or determine the exact decision threshold.  As an example of this procedure, \autoref{app:7} holds the output distributions of a \textbf{uGRU} model with the size "big" for a single sample from the \textbf{XLnL} dataset. Notably, some links have very high consistent confidence, while the prediction varies much more for others.




\begin{figure}
\centering
    \includegraphics[width=0.85\linewidth]{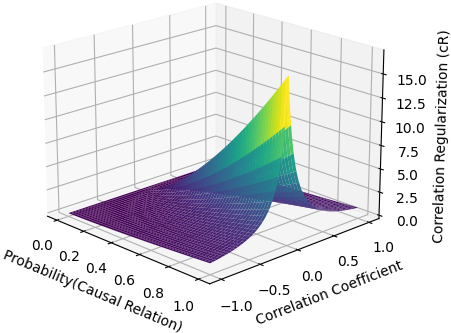}
\caption{Visualization of CR for $\alpha$ = 1.5 and $\beta$ = 0.15. The penalty is only big when the confidence is high while the correlation coefficient is low. }
\label{g:3}
\end{figure}

\begin{table*}[h]
    \centering
    \begin{tabular}{lcc}
    \toprule
        \textbf{Abbreviation} & \textbf{Full Name} & \textbf{Type} \\ 
    \midrule
        CP & Causal Pretraining & Method\\
        CPNN & Causally Pretrained Neural Network & Method\\
        \midrule
        MLP & Multi-Layer Perceptron & Architecture\\
        uGRU & unidirectional Gated Recurrent Unit  & Architecture\\
        bGRU & bidirectional Gated Recurrent Unit  & Architecture\\
        CM & ConvMixer & Architecture\\
        Trf & Transformer  & Architecture\\
        \midrule
        SL & Small Linear & Dataset \\
        ML & Medium Linear & Dataset \\
        SNL & Small Nonlinear & Dataset \\
        MNL & Medium Nonlinear & Dataset \\
        LNL & Large Nonlinear & Dataset \\
        XLNL & Extra Large Nonlinear & Dataset \\

        \midrule
        CP & Correlation Thresholding & Baseline \\
        GVAR & Granger Causal Vectorized Autoregression & Baseline \\
        \midrule
        LCM & Large Causal Model & Model size \\
        \midrule 
        CR & Correlation Regularization & Technique \\
    \bottomrule
    \end{tabular}
    \caption{Summary of abbreviations used throughout this paper.}
\label{app:8}
\end{table*}

\end{document}